# 3D Reconstruction Using a Linear Laser Scanner and A Camera

Rui Wang,

Department of Computer Science and Technology,

Zhejiang University, Zhejiang, China

rainwang6188@gmail.com

*Abstract*—**With the rapid development of computer graphics and vision, several three-dimensional (3D) reconstruction techniques have been proposed and used to obtain the 3D representation of objects in the form of point cloud models, mesh models, and geometric models. The cost of 3D reconstruction is declining due to the maturing of this technology, however, the inexpensive 3D reconstruction scanners on the market may not be able to generate a clear point cloud model as expected. This study systematically reviews some basic types of 3D reconstruction technology and introduces an easy implementation using a linear laser scanner, a camera, and a turntable. The implementation is based on the monovision with laser and has tested several objects like wiki and mug. The accuracy and resolution of the point cloud result are quite satisfying. It turns out everyone can build such a 3D reconstruction system with appropriate procedures.**

##   I.   Introduction *(Heading 1)*

Reconstructing existing objects has been a critical issue in computer vision. After decades of development, 3D reconstruction technology has made great progress. It is currently applied in artificial intelligence, robotics, SLAM (Simultaneous localization and mapping), 3D printing and many areas, which contain enormous potential as well as possibilities[1].

Robert[2] in 1963 first introduced the possibility of acquiring 3D information of an object through 2D approaches. It was then vision based 3D reconstruction became popular and plenty of methods have emerged. In 1995, Kiyasu and his team reconstructed the shape of a specular object from the image of an M-array coded light source observed after reflection by the object[3]. Snavely[4] presented a system for interactively browsing and exploring unstructured collections of photographs of a scene using a novel 3D interface. The advantage of this system is that it can compute the viewpoint of each photograph as well as a sparse 3D model of the scene and image to model correspondences, though the clarity of sparse 3D model is a bit low. Pollefeys[5] and his team presented a system for automatic, geo-registered, real-time 3D reconstruction from video of urban scenes, which uses several images in the video stream to rebuild the scene after feature extraction and the matching of multi-view geometric relations. In 2009, Furukawa et al. proposed a multi-view stereo reconstruction method based on a patch[6]. The advantage of this method is that the reconstructed object has a good body contour completeness, strong adaptability, and does not need to initialize data. In addition, in 2013, the Kinect Fusion project[7] launched by Microsoft Research made a major breakthrough in the field of 3D reconstruction. Unlike 3D point cloud stitching, it mainly uses a Kinect to surround the object and continuously scan it, and the 3D model of the object is reconstructed in real time, which effectively improves the reconstruction accuracy. Microsoft Research announced the Mobile Fusion project[8] at the ISMAR 2015 conference, which uses mobile phones as a 3D scanner and produces 3D images.

Existing 3D reconstruction technologies can be divided into two categories, touch and non-touch, the difference in the technique used to acquire data. Since touch techniques focus on specific conditions and may cause damage to the object, this passage will concentrate on the non-touch category.

As is covered previously, plenty of 3D reconstruction methods have emerged and developed. However, purchasing a 3D reconstruction system with satisfying accuracy still costs a lot. Thus, this passage develops a whole procedure to build an accurate reconstruction system based on monovision with a line laser emitter and a turntable manually. The procedure involves camera calibration, laser plane calibration, rotation axis calibration, scanning object, coordinate computation and point cloud merge. The principle of monovision will be discussed in detail in the article. This implementation indicates that a cheap yet accurate 3D reconstruction system is easy to build by oneself.

##   II.   Techniques for generating point clouds

There are several mature methods to generate point clouds. According to the illuminant is provided or not，these technologies can be divided into two types [9], active vision, and passive vision. The active vision encompasses techniques like laser scanning, structured light, time of flight, and Kinect, while passive vision includes monovision, stereo-vision and multi-vision. The following will give a brief introduction to each of them.

### A. Active Vision

#### 1) Laser Scanning

The laser scanning method uses a laser scanner rangefinder to measure the real scene. First, the laser rangefinder emits a beam of laser to the surface of the object, and then, according to the time difference between the received signal and the transmitted signal, the distance between the object and the laser rangefinder is determined, obtaining the size and shape of the target. The advantage of this method is that it establishes a 3D model of a simple-shaped object, also generates a 3D model of an irregular object, and the generated model is of relatively good accuracy.

*2) Structured Light*

The Structured light method is one of the main directions of research. The principle of the structured light method is to first build a 3D reconstruction system consists of the projection equipment, the image acquisition equipment, and the object to be measured through calibration. Secondly, the surface of the measurement object and the reference plane are respectively projected with a certain regular structured lightmap. Then it uses the vision sensor for image acquisition, so as to obtain the structured light image projection information of the surface of the object to be measured and the reference plane of the object. After that, the acquired image data is processed by triangulation principle, image processing and other technologies, and the depth information of the object surface is calculated, converting the two-dimensional image to 3D image [10][11][12][13].

According to the light pattern used, the structured light method can be divided into point structured light method, line structured light method, surface structured light method, network structured light, and color structured light.

The principle of structured light is shown in Figure 1. The position of the object in the world coordinate $(X_W, Y_W, Z_W)$ has the following relation with the pixel in the image coordinates $(u, v)$ and projection angle $\theta$:

$$[X_W, Y_W, Z_W] = \frac{b}{f\cos\theta - u}[u, v, f] \quad (1)$$

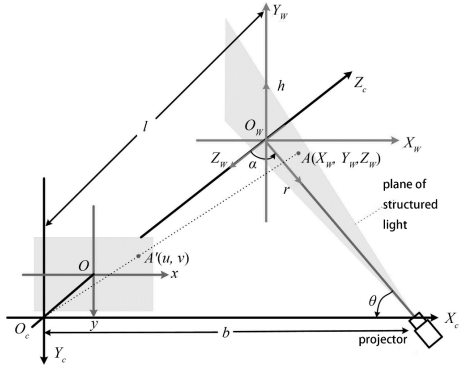

Figure 1. diagram of structured light triangulation

*3) TOF*

TOF (Time of flight) is one of the active ranging technologies which emits the pulsed light and gets the reflected light. This method uses the time difference between emission and reception to determine the distance to the object, so that depth information can be acquired. Its principle is shown in Figure 2.

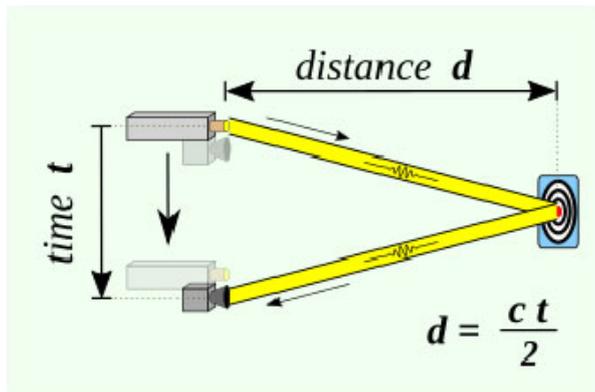

Figure 2. Principle of TOF method

B. *Passicve Vision*

*1) Monovision method*

Monocuar vision is a method that uses only one camera for 3D reconstruction. This method is simple and reliable, and the processing time is relatively short. Thus, this method is widely used in 3D measurement and testing. The implementation of this paper is based on this approach as well.

The principle of Monovision is to convert the 3D world coordinates into 2D image coordinates through a camera. The function of the camera can be abstracted to the product of a camera matrix (intrinsic parameters of the camera, involves scaling) and a rotation matrix (extrinsic parameters, involves rotation and translation). As a result, the process can be illustrated in Figure 3, which is also referred to camera calibration equation.

$$s\begin{bmatrix}u\\v\\1\end{bmatrix} = \begin{bmatrix}f_x & 0 & c_x\\0 & f_y & c_y\\0 & 0 & 1\end{bmatrix}\begin{bmatrix}r_{11} & r_{12} & r_{13} & t_1\\r_{21} & r_{22} & r_{23} & t_2\\r_{31} & r_{32} & r_{33} & t_3\end{bmatrix}\begin{bmatrix}X\\Y\\Z\\1\end{bmatrix}$$

2D Image Coordinates / Intrinsic properties (Optical Centre, scaling) / Extrinsic properties (Camera Rotation and translation) / 3D World Coordinates

Figure 3. Figure 3 of camera calibration

The full process of monovision 3D reconstruction involves monovision video data acquirement, camera calibration, data image process, feature information matching, and 3D information reconstruction.

Monocular vision mainly extracts feature information such as brightness, depth, texture, contour, geometric shape, and feature points in the image, and there is a mass of feature information has been proposed [14], such as SFS (Shape from shading) and PSFS (Perspective shape from shading). As for implementation in this paper, the RGB value of the pixel is used in feature extraction.

*2) Stereo-vision method*

The working principle of stereo-vision 3D reconstruction comes from the human binocular vision system [15], which is the left and right images at the same position are captured by two identical cameras from different perspectives. By employing the triangulation principle, we can obtain the depth information of the object, which helps to reconstruct the 3D model of the object. At present, the 3D reconstruction method based on stereo vision is a potential yet difficult area in the 3D reconstruction technology [16].

Depending on the arrangement of the optical axis, the stereo vision method can be categorized into parallel optical axis and converging optical axis, as is shown in Figure 4.

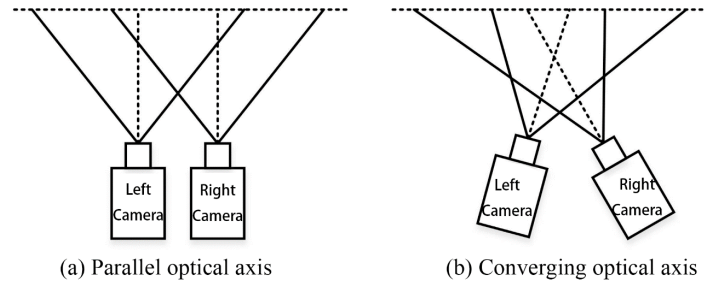

(a) Parallel optical axis        (b) Converging optical axis

Figure 4.    Binocular vision system

Whether parallel or converging, both methods require the steps of camera calibration, image acquirement, alignment correction, and stereo matching, which is similar to the monovision 3D reconstruction.

*3) Multi-vision method*

*Multi-v*ision 3D reconstruction is an extension of binocular vision. By increasing the number of cameras used, it is avoided that some cameras cannot receive reflected light due to the large angle of the object's surface [17].

### III. IMPLEMENTATION

The following will introduce a rough yet accurate 3D reconstruction tool by using only a computer, a line laser scanner, a camera and a turntable, and all systems are built manually (Figure 5).

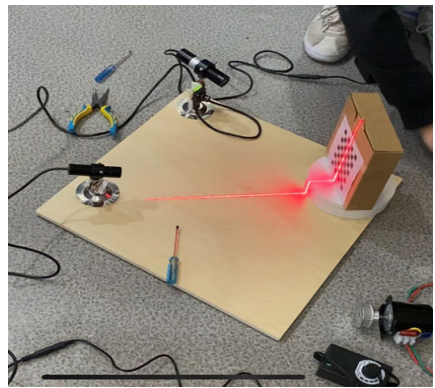

Figure 5.    Overview of 3D reconstruction system

First, based on the flowchart shown in Figure 6, familiarize with the process of using the turntable to perform the 3D reconstruction. It is similar to monovision reconstruction, which is actually the basis of realization, and brightness is selected as the feature. Next, the principle and related code of each process will be introduced.

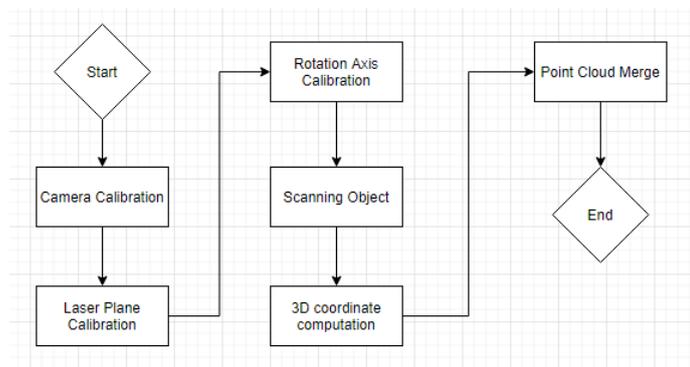

Figure 6.    Flow chart of 3D reconstruction system

#### A. Camera Calibration

In the experiment, defining the real world coordinates with a chessboard is the first step to do. In the process of calibration, one needs to calculate the camera parameters by a set of know 3D points $(X_w, Y_w, Z_w)$ and their corresponding pixel location $(u, v)$ in the image.

To get those 3D points' coordinates, one has to photograph a chessboard pattern with known dimensions at many different orientations. Since the world coordinate is attached to the chessboard and all the corner points lie on the same plane, setting $Z_w$ for every point to be 0 would be a better choice for simplicity. Since points are equally spaced in the chessboard, the $(X_w, Y_w)$ coordinates of each 3D point are easily defined by taking one point as reference $(0, 0)$ and defining remaining with respect to that reference point.

Then capture 20 images of the chessboard from different viewpoints (shown in Figure 7) and find 2D coordinates of those dots using `findChessboardCorners()` and `cornerSubPix()` functions provided by OpenCV (this passage used the OpenCV 3.4.11 version).

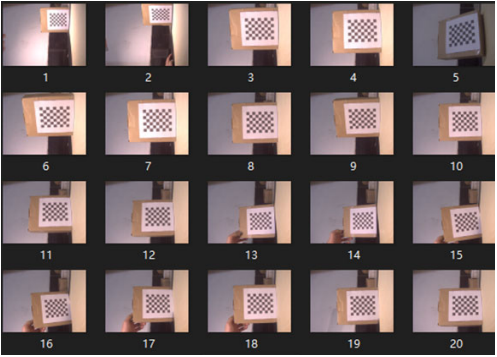

Figure 7. Chessboard from different viewpoint

After that, call openCV's `calibrateCamera()` method to get the parameters like camera matrix, distortion coefficients, rotation vector, and translation vector, the latter 2 parameters can be combined to get the extrinsic parameters.

Since the position of the camera is fixed, the camera coordinate of each picture is the same. Thus, all the operations related to coordination will be performed in the camera coordinate in the following processes. To convert the pixel position $(u, v)$ to camera coordinate, left multiply the inverse of the camera matrix to the pixel coordinate, which can be derived from the equation in Figure 3. The process is illustrated in the formula (2). The Figure 8 is the implemented code.

$$(u, v) \text{ in camera coordinate} = \begin{bmatrix} f_x & 0 & c_x \\ 0 & f_y & c_y \\ 0 & 0 & 1 \end{bmatrix}^{-1} \begin{bmatrix} u \\ v \\ 1 \end{bmatrix} \quad (2)$$

```
1.   cout << "--------------------Step1: Camera Calibration-------------------" << endl;
2.   cv::Mat rgbImage, grayImage;
3.   std::vector<cv::Point2f> corner;
4.   std::vector<std::vector<cv::Point2f>> imagePoint;
5.   for (int i = 1; i <= 20; i++)
6.   {
7.    string path = "./img/calib/" + to_string(i) + ".bmp";
8.    rgbImage = cv::imread(path, CV_LOAD_IMAGE_COLOR);
9.
10.   cout << "Read image success: " << path << endl;
11.
12.   cv::cvtColor(rgbImage, grayImage, CV_BGR2GRAY);
13.
14.   bool isFind;
15.   isFind = findChessboardCorners(grayImage, Size(8, 6), corner, CALIB_CB_ADAPTIVE_THRESH | CALIB_CB_NORMALIZE_IMAGE);
16.   cout << "Find corner!" << endl;
17.
18.   if (isFind)
19.   {
20.    drawChessboardCorners(rgbImage, cv::Size(8, 6), corner, isFind);
```

```
21.        imagePoint.push_back(corner);
22.      }
23.    }
24.
25.    //standard graph are used for projection transformations
26.    std::vector<std::vector<cv::Point3f>> objRealPoint;
27.    calRealPoint(objRealPoint, 8, 6, 20, 3);
28.
29.    //calibration
30.    cv::Mat cameraMatrix, distCoeff;
31.    vector<Mat> rvecsMat;
32.    vector<Mat> tvecsMat;
33.    float rms = calibrateCamera(objRealPoint, imagePoint, cv::Size(rgbImage.cols, rgbImage.rows), cameraMatrix, distCoeff, rvecsMat, tvecsMat, CV_CALIB_FIX_K3);
34.    cout << "Find 20 camera parameter matrix!" << endl;
```

Figure 8.  Code implementation for camera calibration [1]

## B. Laser Plane Calibration

### 1) Laser Line Extraction

Extracting the laser line from the image (i.e., obtaining the pixel coordinates of the laser point) is an important step in calibrating the laser plane. It can be accomplished using the function cv::threshold() .

cv::threshold() perform basic thresholding operations, differentiating the particular pixels from the rest (which will eventually be rejected) by performing a comparison of each pixel intensity value with respect to a threshold value. The extraction effect is shown in Fig 9.

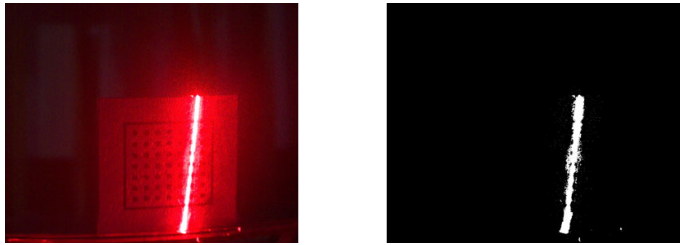

Figure 9.  Example of the laser line thresholding

### 2) Fit Plane

Once the extraction of the pixel coordinates of the laser line in the image is finished, the method mentioned in formula (2) is used to transform it to the camera coordinate.

The problem can be rephrased as fitting the plane $ax + by + cz = d$ with n different points $(x_i, y_i, z_i)$.

The SVD (singular value decomposition) is used to perform the fit plane. It is performed as follows:

$$ax + by + cz = d \quad (3)$$

$$\text{suppose } (\bar{x}, \bar{y}, \bar{z}) \text{ is the mean position, then } a\bar{x} + b\bar{y} + c\bar{z} = d \quad (4)$$

$$a(x_i - \bar{x}) + b(y_i - \bar{y}) + c(z_i - \bar{z}) = 0 \quad (5)$$

$$\text{Let } A = \begin{bmatrix} x_1 - \bar{x} & y_1 - \bar{y} & z_1 - \bar{z} \\ x_2 - \bar{x} & y_2 - \bar{y} & z_2 - \bar{z} \\ \ldots & \ldots & \ldots \\ x_n - \bar{x} & y_n - \bar{y} & z_n - \bar{z} \end{bmatrix}, X = \begin{bmatrix} a \\ b \\ c \end{bmatrix} \quad (6)$$

$$\text{Then we have } AX = 0$$

The eigenvector of the smallest singular value of A just corresponds to the plane parameters a, b, c, d

Figure 10 is the point cloud of two calibrated laser line generated in the experiment. It is obvious that they're on the same plane. The code implementation is shown in Figure 11.

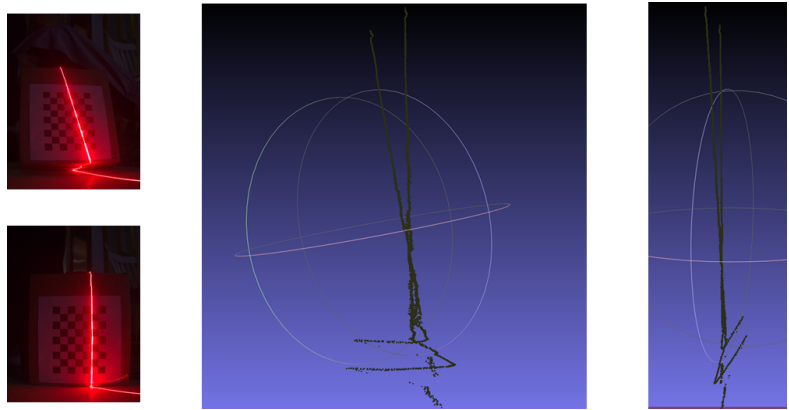

Figure 10. Example of two laser line point cloud

```
1.   //Plane equation: Ax+by+cz=D
2.   void cvFitPlane(const CvMat* points, float* plane) {
3.    // Estimate geometric centroid.
4.    int nrows = points->rows;
5.    int ncols = points->cols;
6.    int type = points->type;
7.    CvMat* centroid = cvCreateMat(1, ncols, type);
8.    cvSet(centroid, cvScalar(0));
9.    for (int c = 0; c < ncols; c++) {
10.    for (int r = 0; r < nrows; r++)
11.    {
12.     centroid->data.fl[c] += points->data.fl[ncols * r + c];
13.    }
14.    centroid->data.fl[c] /= nrows;
15.   }
16.   // Subtract geometric centroid from each point.
17.   CvMat* points2 = cvCreateMat(nrows, ncols, type);
18.   for (int r = 0; r < nrows; r++)
19.    for (int c = 0; c < ncols; c++)
20.     points2->data.fl[ncols * r + c] = points->data.fl[ncols * r + c] - centroid->data.fl[c];
21.   // Evaluate SVD of covariance matrix.
22.   CvMat* A = cvCreateMat(ncols, ncols, type);
23.   CvMat* W = cvCreateMat(ncols, ncols, type);
24.   CvMat* V = cvCreateMat(ncols, ncols, type);
25.   cvGEMM(points2, points, 1, NULL, 0, A, CV_GEMM_A_T);
26.   cvSVD(A, W, NULL, V, CV_SVD_V_T);
27.   // Assign plane coefficients by singular vector corresponding to smallest singular value.
28.   plane[ncols] = 0;
29.   for (int c = 0; c < ncols; c++) {
30.    plane[c] = V->data.fl[ncols * (ncols - 1) + c];
31.    plane[ncols] += plane[c] * centroid->data.fl[c];
```

```
32.     }
33.     // Release allocated resources.
34.     cvReleaseMat(¢roid);
35.     cvReleaseMat(&points2);
36.     cvReleaseMat(&A);
37.     cvReleaseMat(&W);
38.     cvReleaseMat(&V);
39. }
```

Figure 11.  code implementation for fitting plane [1]

*C. Rotation Axis Calibration*

To calibrate the rotation axis, one should position the checkerboard to pass through the center of the turntable, and then hit a vertical laser beam over it to use this infrared ray as the axis of rotation. Figure 12 is a photo taken during the axis calibration.

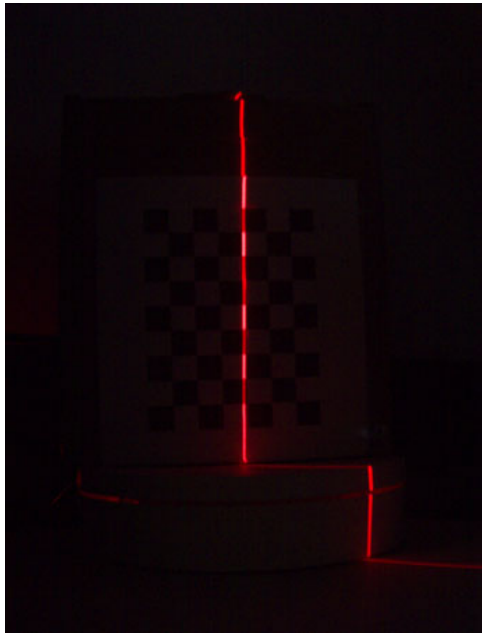

Figure 12. Calibration the rotation axis

By extracting the particular laser line which is assumed as a rotation axis and transform it into a camera coordinate, it can perform line fit to obtain the line equation of the rotation axis.

*D. Scanning Object*

When all the calibration work is done, the reconstruction system is capable of scanning the object using a laser line projector. While the object is rotating on the turntable and being projected by a laser line, the camera takes the whole process.

*E. 3D coordinate computation*

In this process, first, the laser line needs to be extracted from the scanning step, and the points in the image coordinates are obtained. Second, transform them to the camera coordinate via formula (2). Now all the points of each line in 3D coordinate are acquired, the next step just needs to reverse the rotate process to merge all those points to get the object model.

*F. Point Cloud Merge*

It is noticed that the object is performing a 360 degrees rotation. As a result, the merging problem can be abstract to performing a rotation around an arbitrary axis with all the points on a curve. It can be described in the following steps, as shown in Figure 13. First, it transforms and rotation so that the center axis coincides with the X-axis. Then perform the rotation of those points since the axis is the coordinate axis. Lastly, apply the inverse transform of the first step so that center axis is back to the original position.

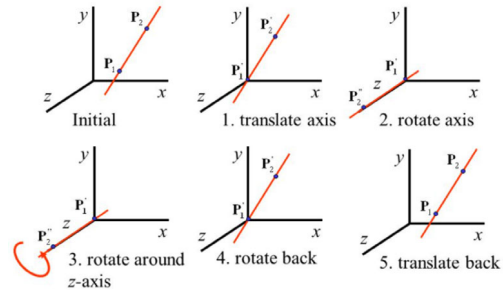

Figure 13.  Procedures of rotating a point around an arbitrary axis

IV. RESULT

This passage tested the 3D reconstruction system with several objects, including a wiki (Figure 14), a mug (Figure 15) and a plush toy (Figure 16). For a better view of the 3D point cloud model, view directly in ply file in the github repository [1]. It turns out that the point cloud generated is quite satisfying.

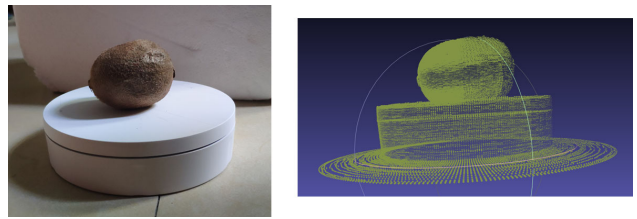

Figure 14.  The original wiki and the reconstructed wiki [1]

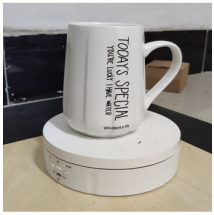 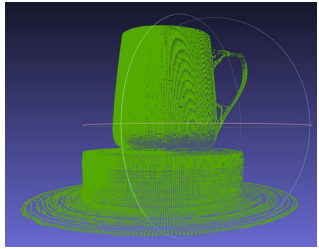

Figure 15. The original mug and the reconstructed mug [1]

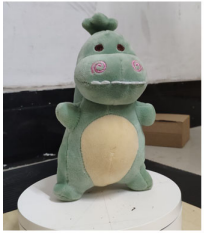 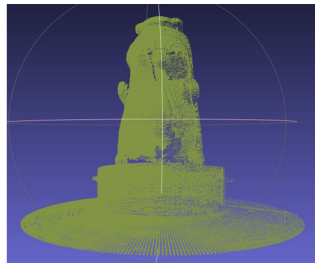

Figure 16. The original toy and the reconstructed toy [1]

## V. CONCLUSION

The quality of the generated point cloud basically relies on the accuracy of the calibration. The more accurate the calibration process is carried out, the clearer the point cloud model would be. Another fact that contributes to the quality of the reconstructed model is the object, especially its texture. Comparing the mug model in Figure15 with the toy model in Figure 16, it's obvious that the mug model is clearer as well as accurate. Since the dragon toy is furry, the diffusion of the laser light makes the laser line extraction process more difficult, and there may have some errors.

Though the theoretical knowledge of monovision 3D reconstruction is not that difficult, building it from scratch is still challenging and frustrating. The pivotal process is the calibration, and it significantly influences the accuracy of the final point cloud model. It took a lot of time and the dot calibration gird is used as well to perform the calibration. Fortunately, the result is quite satisfying.

The implemented 3D reconstruction system based on monovision and laser line produces quite satisfying point cloud models. As the cheap cost as well as uncomplex building procedure, it is convenient for anyone to rebuild.

In the future, encapsulating the whole system in a portable box is needed, allowing users to reconstruct objects anywhere conveniently. Moreover, this research will further improve the scanning and extraction method as well, so that this system will have better accuracy of reconstructed point cloud models of fluffy objects.